\documentclass[letterpaper, 10 pt, conference]{ieeeconf}  

\IEEEoverridecommandlockouts                              

\usepackage{amsmath} 
\usepackage{amssymb}  

\usepackage{cite}
\usepackage{caption}
\usepackage{subcaption}
\usepackage{graphicx}
\usepackage{graphbox}
\usepackage{wrapfig}
\usepackage{float}
\let\labelindent\relax 
\usepackage{enumitem}
\usepackage{textcomp}
\usepackage{svg}
\usepackage{multirow}
\usepackage{booktabs}       
\usepackage{hyperref}

\usepackage{diagbox}
\usepackage[export]{adjustbox}

\title{\LARGE \bf
Merging Decision Transformers: \\Weight Averaging for Forming Multi-Task Policies
}

\author{Daniel Lawson and Ahmed H. Qureshi\thanks{The authors are with the Department of Computer Science, Purdue University, West Lafayette, IN 47907 USA (email: lawson95@purdue.edu; ahqureshi@purdue.edu)}}

\begin{document}

\maketitle
\thispagestyle{empty}
\pagestyle{empty}

\begin{abstract}

Recent work has shown the promise of creating generalist, transformer-based, models for language, vision, and sequential decision-making problems. To create such models, we generally require centralized training objectives, data, and compute. It is of interest if we can more flexibly create generalist policies by merging together multiple, task-specific, individually trained policies. In this work, we take a preliminary step in this direction through merging, or averaging, subsets of Decision Transformers in parameter space trained on different MuJoCo locomotion problems, forming multi-task models without centralized training. We also demonstrate the importance of various methodological choices when merging policies, such as utilizing common pre-trained initializations, increasing model capacity, and utilizing Fisher information for weighting parameter importance. In general, we believe research in this direction could help democratize and distribute the process that forms multi-task robotics policies. Our implementation is available at \href{https://github.com/daniellawson9999/merging-decision-transformers}{https://github.com/daniellawson9999/merging-decision-transformer}. 
\end{abstract}

\section{INTRODUCTION}

Following advances in sequence modeling, there have been various works studying the feasibility of creating generalist transformer-based policies across many control tasks, with work such as Multi-game Decision Transformer \cite{multi-game-dt} creating multi-task policies for Atari, or Gato \cite{Gato} which learns a single generalist across many modalities, tasks, and robot embodiments. We consider the nature of which these multi-task robotics policies can be formed. Is it possible to more flexibly create multi-task control policies with reduced demands for centralizing all data and training simultaneously? 

We believe an initial step in this direction is investigating the feasibility of combining single Decision Transformers (DT) trained on different tasks. Specifically, we consider DTs trained on MuJoCo locomotion problems. In this work, we look at combining through the perspective of weight merging \cite{fisher-merging, modelsoup, choshen2022fusing}. For example, given two trained DTs on different environments, merging refers to taking all parameters or a subset of parameters, say those associated with attention, and replacing the parameters in both models with combinations of their original parameters.

\begin{figure}[t]
    \centering
    \begin{minipage}{\linewidth}
    \vspace{0px}
    \begin{subfigure}{\textwidth}
    \includegraphics[width=.975\linewidth]{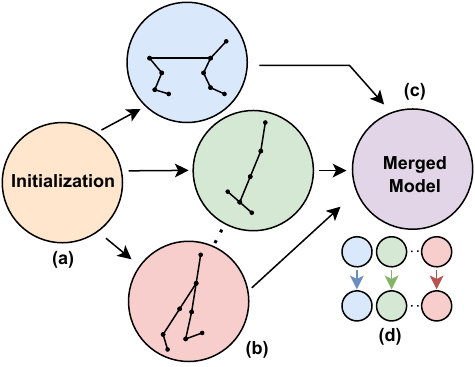}
    \end{subfigure}
    \end{minipage}\hfill
    \begin{minipage}[b]{\linewidth}
        \caption{\textbf{Process for merging Decision Transformers}. Starting from common, or unique initializations (a), we separately train models for different offline reinforcement problems (b). After training, task-specific models are combined by merging over a subset of parameters (c). After merging, we may improve performance by freezing merged parameters, and then separately finetuning, updating unique un-merged parameters (\textit{Merge-Freeze-Finetune}) (d).\\}
    \label{fig:dt1}
    \end{minipage}
    \vspace{-46px}
\end{figure}

In this work, we assess the feasibility of merging transformers for control tasks, and provide several techniques to achieve successful merging, forming our contributions:
\begin{enumerate}[leftmargin=*]
    \item We begin by investigating how merging individual layers and subsets of DTs affect model performance and find that we can directly merge and swap certain parameters of Decision Transformers trained on different MuJoCo environments (HalfCheetah, Walker, Hopper) with, in some cases, minimal decrease in performance. This leads us to investigate the role of attention in DTs, finding some DTs do not heavily rely on attention.  
    \item We propose a method for creating multi-task DTs through merging certain parameters, freezing those merged, and then independently finetuning un-merged parts with \textit{Merge-Freeze-Finetune} (MFF). This creates a multi-task DT without centralized data or training objectives. We also demonstrate improving existing sub-optimal multi-task policies through merging without MFF of finetuned policies on higher quality data.      
    \item We show that common initialization can lead to better performance after merging with Decision Transformers. Specifically, we use common initializations from language modeling \cite{wiki, modalities}. We also investigate methods to improve merging, such as co-training and regularization to model initializations, increasing model size, and utilizing Fisher information. 
\end{enumerate}

\section{Background}
\subsection{Transformers}
\label{sec:transformers}
Transformers \cite{allyouneed} are a common neural network architecture for modeling sequences. We consider causal decoder-only transformers like GPT \cite{GPT}. A transformer is composed of several successive transformer blocks, which each consist of (multi-headed) self-attention layers, multi-layer perceptron (MLP) layers, as well as layer normalization \cite{layernorm}. Given a sequence of inputs of length $n$ and embedding size $d$ $X \in \mathbb{R}^{d\times n}$, a self-attention layer projects each input to queries ($Q$), keys ($K$), and values ($V$) with parameters $\{W_q, b_q\},\{W_k, b_k\},\{W_v, b_v\}$ respectively. We then perform Attention($X$):
$\mathrm{softmax}(\frac{QK^T}{\sqrt{d}})V
$. We apply this operation for each head in parallel, stack outputs, resulting in $Y \in \mathbb{R}^{Hd \times n}$, where $H$ is the number of heads, which is then projected by $W_o$, resulting in $X' = W_oY \in \mathbb{R}^{d \times n}$ \cite{formaltransformer}. Decision Transformers, apply layer normalization before and after attention layers, where each layer has learnable affine transform parameters $\gamma, \beta$. Each MLP contains one hidden layer (and one output layer), with parameters $W_1, W_2$. Transformers also have a residual connection after each attention and MLP layer respectively, so the actual output consists of the transformation from the layer added with its input.  

\subsection{Offline reinforcement learning with decision transformers}
We consider problems that are modeled by a Markov decision process (MDP) with states ${s} \in \mathcal{S}$, actions ${a} \in \mathcal{A}$, unknown transition dynamics $p(s'|s,a)$ and reward function $r(s,a)$. Traditionally in RL, we aim to learn an optimal policy that maximizes expected (discounted) return through interaction with the environment. Instead, in offline RL, we learn without interaction using a static dataset. A dataset consists of a set of trajectories, where each trajectory has the form $\tau = ({s}_{0},{a}_{0},r_{0},{s}_{1},{a}_{1},r_{1},...,{s}_{N},{a}_{N},r_{N})$, consisting of a sequence of states, actions, and return at each timestep until timestep $N$. With this data, we aim to find a policy $\pi(a | \cdot)$ that maximizes expected return $\mathbb{E}[\sum_{t=0}^{N}r_{t}]$, where the expectation is over the distribution induced by transition dynamics and the policy $\pi$. Decision Transformer \cite{dt} casts offline RL as a sequence modeling problem using causal auto-regressive transformers. Particularly, DT models the actions in the sequence $\tau=(\hat{R}_{1},s_{1},a_{1},\hat{R}_{2},s_{2},a_{2},\dots,\hat{R}_{T},s_{T},a_{T})$, where the return-to-go (RTG) is the undiscounted sum of future reward: $\hat{R}_{t}=\sum_{t^{\prime}=t}^{T}r_{t^{\prime}}$. To make a prediction, each input is embedded using linear projections with added positional encoding and normalization, passed through the transformer, and then a final linear layer predicts the action. We refer to transformer parameters as those only associated with transformer layers and not input or output projections. 


\section{Methods}
In this work, we consider the problem of merging or combining neural networks trained on independent robotics control tasks. Given $m$ networks $\{\theta_1, \dots,\theta_m\}$, we aim to obtain a merged network $\theta_M$ which combines the weights of each individual network, obtaining a new model which satisfies all individual task but with a similar number of parameters as just a single model. 

To obtain multi-task models, we investigate both direct averaging and fisher merging, which do not require using the prior data, or performing additional training. We demonstrate the success of merging without additional training after the merging step for improving sub-optimal multi-task models in Section \ref{sec:ext}.

When merging single-task models trained on separate robotic embodiments, we may encounter interference when merging. If we relax the constraint of performing additional training, and having access to prior data, we study whether we can still form a multi-task model without centralized training by separately finetuning only a few parameters for each task after merging.

\subsection{Merging}
Given two trained neural networks with the same architecture, and their sets of parameters or weights $\theta_A, \theta_B$, we can interpolate between weights, getting a new model $\theta_\lambda = (1 - \lambda)\theta_A + \lambda\theta_B$. With $\lambda = .5$, we directly average weights, obtaining a merged model $\theta_{\lambda = .5} = \theta_M = .5(\theta_A + \theta_B)$. In this work, we are interested in merging subsets of transformer parameters discussed in Section \ref{sec:transformers}.

 Directly interpolating between parameters of randomly initialized models may result in models with high loss \cite{roleofperm, basin}, but this can be mitigated instead merging models from common initializations. When models $\theta_A, \theta_B$ are  from two finetuned models from a common initialization, $\theta_{\mathrm{pre}}$, merging is equivalent to task arithmetic for creating multi-task models \cite{taskarithmetic}. Specifically, a task vector $\tau_x = \theta_x - \theta_{\mathrm{pre}}$ stores the difference between a finetuned and pre-trained model. We can form a multi-task model by adding the averaged task vectors to the original model: 

 \begin{equation}
 \begin{split}
\theta_{\mathrm{pre}} + \frac{1}{2}(\tau_A + \tau_B) = \theta_{\mathrm{pre}} + \frac{1}{2}((\theta_A - \theta_{\mathrm{pre}}) \\+ (\theta_B - \theta_{\mathrm{pre}})) = \frac{1}{2}(\theta_A + \theta_B) = \theta_M \end{split}\end{equation}

 \begin{figure*}[h]
\centering
\begin{subfigure}{.44\textwidth}
  \centering
    \includegraphics[width=1\textwidth, trim={0 0 6cm 0},clip]{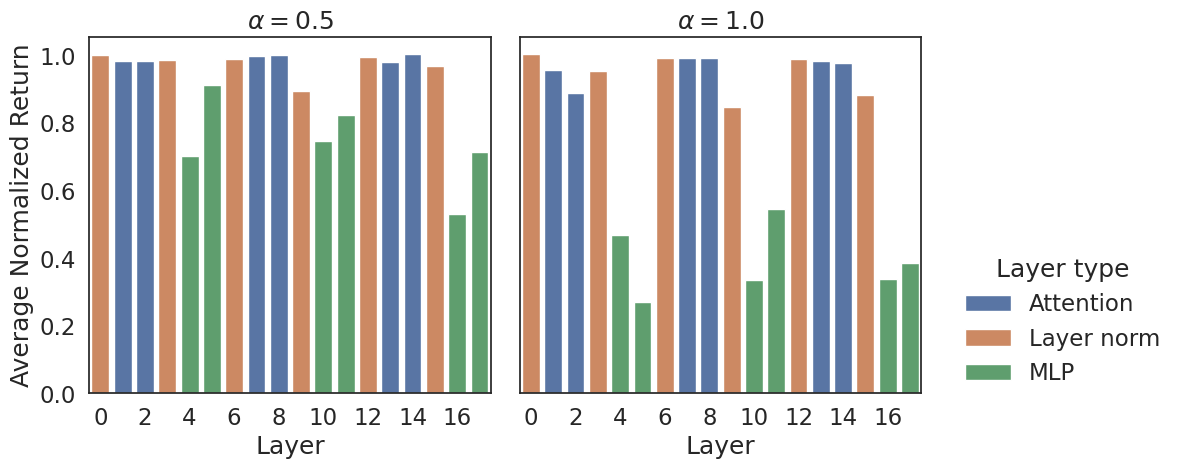}
  \caption{}
  \label{fig:dt1}
\end{subfigure}%
\begin{subfigure}{.45\textwidth}
  \centering
    \includegraphics[width=1\textwidth, trim={0cm 0 6cm 0},clip]{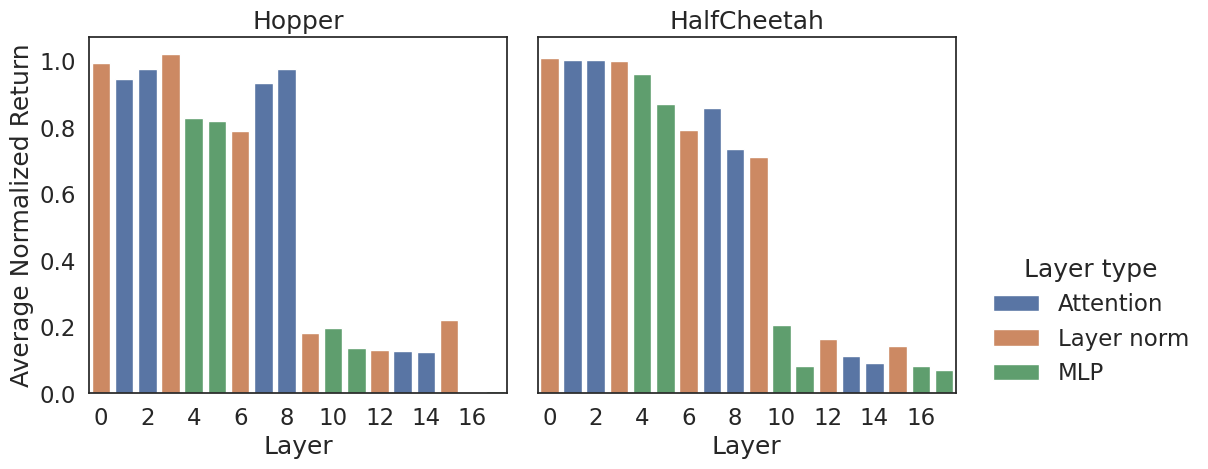}
    \caption{}
    \label{fig:incremental}

\end{subfigure}
\begin{subfigure}{.10\textwidth}
  \centering
    \includegraphics[width=1\textwidth, trim={25cm -2cm 0cm 0},clip]{images/mujoco/layer/hc-inc-naive5-l.png}
\end{subfigure}

\caption{(a): Normalized return after \textbf{merging a single layer} at a time, averaged between all pairs of Walker, Hopper, HalfCheetah. Return is normalized by original performance before merging. We evaluate at $\alpha=.5$. an average, and $\alpha=1$, swapping in a layer from another DT. (b): \textbf{Incremental merging} with $\alpha=.5$ merging from HalfCheetah to Hopper (\textbf{left}) and Hopper to HalfCheetah (\textbf{right}), where the set of merged layers grows as moving further in depth.}
\label{fig:test}
\end{figure*}
 
\subsubsection*{Fisher merging}
Another interpretation of merging $M$ models which start from a common initialization is that our goal is to finding the parameters $\theta_M$ which maximize the likelihoods of the posterior distribution under all models. If each posterior distribution over parameters is an isotropic Gaussian, then the merged parameters which maximize the joint posterior are: \begin{equation}
\theta_M = \mathrm{argmax}_\theta\sum_i\mathrm{log}p(\theta|\theta_i, I) = \frac{1}{M}\sum_{i}{\theta_i}\end{equation} 

Instead, Fisher merging \cite{fisher-merging} considers a Gaussian approximation $p(\theta|\theta_i, F_i$), where $F_i$ is a diagonal precision matrix containing the Fisher information for each parameter:

\begin{equation}F_i = \mathbb{E}_{x\sim D_i}\mathbb{E}_{y\sim p_{\theta_i}(y|x_i)}\nabla_{\theta_i}(\mathrm{log}~p_{\theta_i}(y|x))^2\end{equation}



For each parameter $j$ of model $i$, this obtains a score $F_i^j$ which represents the precision of the estimate for $\theta_i^j$. This can be computed after training, and stored along with a model. To form a merged model, we can perform a weighted average of model parameters in proportion to each model's Fisher information: \begin{equation}\theta_M^j=\frac{\sum_{i=1}^{M}F_{i}^{j}\theta_{i}^{j}}{\sum_{i=1}^{M}F_{i}^{j}}\end{equation}
As models are trained for different robotic tasks, we find that the scale of gradients vary, so we normalize each parameter's Fisher information by the sum of total information for that model before averaging. 

\subsection{Merge-Freeze-Finetune}
In some instances, we expect that we can directly merge the weights of several models, but in other situations where models are significantly different, we expect significant decreases in performance after merging. In these instances, we propose \textit{Merge-Freeze-Finetune} (MFF): Merging over a subset of two networks or more networks, freezing parameters in the subset, and then separately finetuning the transformer un-merged parameters and linear input and output projections in both original models, to separately adapt to changes in shared parameters.

\subsection{Pre-trained Initializations}

Prior work has shown the importance of common-initialization for merging models trained on different data distributions \cite{fisher-merging, taskarithmetic}. In domains like language modeling, we can simply merge finetuned models from common language model initializations, but we do not have readily available equivalent initializations in robotics. However, we propose that language model initializations can be useful for merging models even in a drastically different domains such as robotics control tasks. This is motivated by work which shows that transformers pre-trained on language can learn general representations and parameters that are amenable to transfer \cite{universalcomp}. In the context of offline reinforcement learning, \cite{wiki, modalities} showed that initializing Decision Transformers (DT) \cite{dt, genDT} with pre-trained language models can increase convergence speed and performance on the D4RL \cite{D4RL} MuJoCo \cite{mujoco} benchmarks, showing transfer between completely different modalities.

Following this motivation, later, in Section \ref{sec:lm} we pre-train with language, utilizing a small LM, ChibiT \cite{wiki} pre-trained on Wikitext-103 \cite{wikitext} with the same architecture we next discuss in Section \ref{sec:exp}. When training a DT with ChibiT initialization, we use the modified objective: \begin{equation}\mathcal{L(\theta)} = \mathcal{L}_{\mathrm{MSE}} + \lambda_1\mathcal{L}_{\mathrm{cos}} + \lambda_2\mathcal{L}_{\mathrm{LM}} + \lambda_3\mathcal{L}_{\mathrm{2}}
\label{eq:chibi}
\end{equation}
\quad Where $ \mathcal{L}_{\mathrm{MSE}}$ is the original DT objective, $\mathcal{L}_{\mathrm{LM}}$ is the language modeling objective, $\mathcal{L}_{\mathrm{cos}}$ encourages cosine similarity between DT input embeddings and clustered centers of language token embeddings, and $\mathcal{L}_{\mathrm{2}} = \|\theta_{\mathrm{LM}}-\theta\|_2$.





\section{Related work}
Several papers have studied merging models with shared initialization in language and vision for the purpose of improving performance, generalization, forming pre-trained models, or for extending task capabilities \cite{choshen2022fusing, modelsoup, branchtrainmerge, ilharco2022patching, taskarithmetic, adaptersoup, datalessfusionmerging}, increasing robustness to distribution shift \cite{robust}, and for distributed training \cite{fedavg, coldfusion}.  Other work has improved methods for merging weights by utilizing Fisher information for weighted averaging \cite{kirkpatrick2017overcoming, fisher-merging}, and accounting for symmetries, such as permutation \cite{basin, jordan2022repair, sinkhorn} which is used to effectively merge weights found through optimizing the same loss but with different initialization. To our knowledge, we are the first work that considers merging for decision-making or robotic control settings, which presents unique challenges. Many recent works have considered creating multi-task or general models for decision-making problems \cite{multi-game-dt, multi-q, Gato, VIMA-manipulation, universal-policies}, but rely on simultaneous and centralized training over all tasks. While we focus on experiments framed as multi-task problems, merging, could be explored for continual (reinforcement) learning \cite{continual-rl}, which has long-standing goals of creating systems that can continuously adapt to new tasks. 


\section{Experiments}
\label{sec:exp}

We begin by partially merging Decision Transformers to investigate if similar parameters may be learned across environments, ignoring input and output projection layers which have unique dimensionality and function for different environments.. We use settings for MuJoCo experiments from \cite{dt, wiki}, as well as their implementations. Following their configuration, we use transformers with an embedding size of 128, 1 head, and 3 layers. We randomly initialize and train a model for HalfCheetah, Hopper, and Walker2D on expert D4RL \cite{D4RL} datasets. Given two trained models on different environments, we call the model which is having a layer altered the target model with parameters $\theta_t$, and the model where the layer is taken from, which is trained on another environment, the source model with parameters $\theta_s$. 

We first average a single layer at a time, keeping all other layers unaltered. For one layer $\theta^i \in \theta$, we use the update $\theta_t^{i} = (1-\alpha)\theta^i_t + \alpha(\theta^i_s)$, where $\alpha \in \{.5,1\}$, relating to averaging and swapping parameters. We merge between each pair of models out of the three models covering each environment. For each pair, we look at merging in each direction, swapping source and target models. This leads to six  evaluations per layer, which we average and display in Figure \ref{fig:dt1}. We report normalized return over 25 episodes, where $1$ corresponds to the original return before any merging ($\alpha = 0$). 


 We find that we can both average ($\alpha = .5$), or directly use a layer from another transformer ($\alpha = 1$), showing that Decision Transformers trained on different MuJoCo tasks may learn functionally similar parameters. We see a larger drop-off from merging parameters within the multi-layer perception (MLP) layers. However, we find that we can directly use the attention parameters from another trained model at any depth with little to no decrease in empirical return. 
 


\subsection{Merging subsets of multiple layers}

We should draw attention to that in Figure \ref{fig:dt1} we are merge one layer at a time, and possibly, a transformer could be robust to single-layer interference. We must see how well merging performs with multiple layers at a time, as errors could compound. We look at this from two perspectives. We begin by following a similar procedure to the former per-layer merging, but instead incrementally add an additional layer, forming a larger subset as we go move further in depth, in Figure \ref{fig:incremental}. Instead of averaging across different environment pairs, we show this within a single pair to be able to clearly see the effects of adding layers. We see that as we add layers, performance decreases, reaching close to zero performance when merging the entire transformer. We also tend to see large drop-offs at MLP layers, but less so at attention layers, and varying results with layer normalization. 

\begin{figure}[h]
  \begin{center}
    \includegraphics[width=1\linewidth]{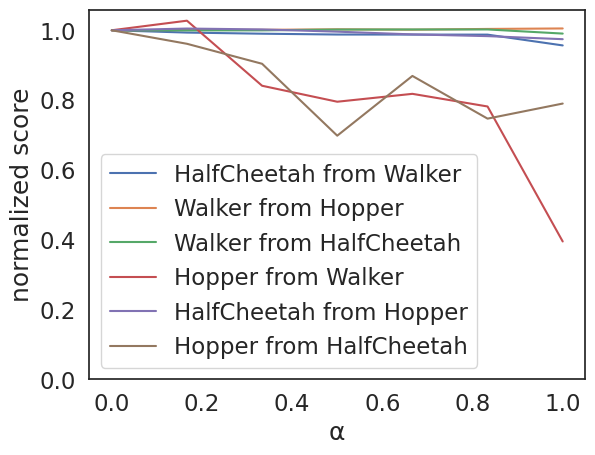}
  \end{center}
  \caption{\textbf{Merging all attention parameters} of transformer across pairs of MuJoCo environments. We vary $\alpha$, which interpolates between original parameters ($\alpha=0$) and to parameters from the second environment ($\alpha=1$).}
  \label{fig:attn}
   \vspace{-10px}
\end{figure}

\subsubsection{Merging all attention layers}
\label{sec:all-attn}

When merging individual layers, we see the least reduction in performance with attention layers. Thus, we see what happens when merging all attention parameters (the parameters associated with the query, key, value, and output projections) at once. Attention parameters consist of $\sim33.18\%$ of the 3 transformer layers which have a total size of $596\mathrm{K}$ parameters. We display results in Figure \ref{fig:attn}. On the x-axis, we interpolate between parameters where $\alpha=0$ corresponds to using unchanged attention parameters and $\alpha=1$ corresponds to using all attention parameters from the other environment. It appears that Decision Transformers trained on different MuJoCo can learn functionally similar attention weights. We can directly swap in the weights of HalfCheetah into the Walker2D transformer, and vice versa, with no decrease in return. We do not see this hold as strongly between Hopper/Walker2d, and Hopper/HalfCheetah, but we still see good results. 

\subsection{Analysing attention merging} 

To explain the success of swapping attention parameters in the prior section, we perturb attention parameters after training to see how much randomly initialized DTs rely on the attention mechanism. We replace the attention weights of trained DTs using both random parameters (fixed for all experiments), identity parameters (weights set to 1, and biases equal to 0), and removing attention, having information pass through residual connections. We show the results of these variations in Table \ref{table:attention}. We see varied results depending on the environment and dataset pair. We see decreases over all datasets with Hopper. With Walker and HalfCheetah, while we see little impact on the homogeneous medium or expert datasets, we see performance decrease after perturbation for models trained on medium-expert. 

\begin{table}[h]
  \small
  \centering
    \caption{We report the impact of \textbf{altering attention parameters of trained DTs}. We report D4RL normalized scores of original performance, replacing attention parameters with randomly initialized parameters, identity parameters, and removing attention, relying on residual connections.}
  \label{table:attention}

  \begin{tabular}{clcccc}
    \toprule
    Dataset     & Env & Original & Random  & Identity  & Removed \\
    \midrule
    \multirow{3}{*}{\shortstack{Medium\\Expert}} 
    & Cheetah & $88.6$ & $41.0$ & $41.2$ & $41.5$ \\ 
    & Hopper & $108.3$ &$ 77.2$ & $66.8$ & $74.1$ \\ 
    & Walker & $109.4$ & $84.6$ & $82.6$ & $86.9$ \\ 
    \midrule
    \multirow{3}{*}{Medium} 
    & Cheetah & $40.4$ &$ 40.3$ & $39.7$ & $40.8$ \\ 
    & Hopper & $76.4 $& $52.2$ & $52.6$ & $52.2$ \\ 
    & Walker & $75.3$ & $74.1$ & $76.2$ & $74.3$ \\ 
    \midrule
    \multirow{3}{*}{Expert}
    & Cheetah & $89.9$ & $90.1$ & $89.9$ & $90.1$ \\ 
    & Hopper & $108.8$ & $67.1$ & $67.9$ & $68.8$ \\ 
    & Walker & $109.4$ & $109.7$ & $109.8$ & $109.8$ \\ 
    
    \bottomrule
  \end{tabular}
\end{table}

\subsection{Merge-Freeze-Finetune}
\label{sec:mff}
In previous sections, even when merging over a compatible subset, such as attention layers, we still see some drop-off in return after merging (such as between Hopper and HalfCheetah), as shown in Figure \ref{fig:attn}. We aim to see if we can merge without losing performance and over parameters which previously saw large decreases in performance, such as MLP parameters as in Figure \ref{fig:dt1}. To overcome this, we test \textit{Merge-Freeze-Finetune} on several subsets, reporting the performance of the merged model as a percent of the original performance. We also report the size increase of the multi-task model within transformer layers, which does not include linear input/output projections that are kept unique for each task. For example, without any merging, we maintain original performance ($100\%$ on each task) but need unique transformer layers for each task ($200\%$). We use subsets of all attention related parameters, MLP and attention parameters, and the entire transformer which also adds layer normalization layers. We also report attention merging without finetuning (M), as previously shown in Section \ref{sec:all-attn}, and an alternative to equally merging, where we keep one model unmodified, but copy and freeze its parameters to the DT in the second environment, and have this model update its non-transformer parameters.

\begingroup
\setlength{\tabcolsep}{4pt} 

\begin{table}[h]
\small
      \centering
      \caption{Evaluating performance after \textbf{merging between HalfCheetah and Hopper expert models} with \textit{Merge-Freeze-Finetune} (MFF) and just merging (M), over different model subsets.}
  \label{table:mff}

    \begin{tabular}{clll}
        \toprule    
        Configuration & \shortstack{Hopper} & \shortstack{Cheetah} & Transformer Size  \\ 
        \midrule
        Original & $100\%$ & $100\%$ & $200\%$  \\ 
        \shortstack{Frozen from Hopper} & $100\%$ & $3\%$ & $100\%$  \\
        \shortstack{Frozen from HalfCheetah} &$48\%$ &$100\%$ & $100\%$  \\ 
        Transformer (MFF) & $61\%$ & $86\%$ & $100\%$  \\ 
        Attention+MLP (MFF) & $90$\% & $98\%$ & $100.26\%$  \\ 
        Attention (MFF) & $101\%$ & $101$\% & $166.82$\%  \\ 
        Attention (M) & $79$\% & $98$\% & $166.82\%$ \\ 
        \bottomrule
    \end{tabular}
\end{table}
\endgroup
We report results between merging Cheetah and Hopper models in Table \ref{table:mff}. When merging over attention layers, plus finetuning (MFF), we can recover the original performance. We can also retain greater compression with Attention+MLP merging, obtaining a multi-task model, with only 1.0026 times more transformer parameters, but with some reduced performance. 

\begin{table*}[h]
  \centering
   \small
    \caption{\textbf{Merging Decision Transformers trained with shared language model initialization} on D4RL medium and medium-expert datasets. We evaluate merging attention + MLP layers with \textit{Merge-Freeze-Finetune} (MFF) over all three environments at once. We report mean D4RL normalized scores and standard error over two instances of MFF, and two multi-task models. Single-task results are reported from DT \cite{dt}.}
  \label{table:lm-merge}
  \begin{tabular}{clllll|l@{\hskip .09in}l}
    \toprule
    \textbf{Dataset}     &  \textbf{Environment} & \textbf{ChibiT} & \textbf{ChibiT$_\mathcal{CO}$}  & \textbf{ChibiT$_\mathcal{R}$} & \textbf{Random} & \textbf{Multi} & \textbf{Single} \\
    \midrule
    \multirow{3}{*}{\shortstack{Medium}} 
        & HalfCheetah & $39.9 \pm 0.3$ & $39.9 \pm 0.3 $ & $41.2\pm0.2$ & $33.8\pm0.5$ & $38.7\pm 0.4$ & $42.6$ \\ 
        & Hopper & $59.4 \pm 0.9$ & $55.4 \pm 0.4$ & $56.8 \pm 0.4$ & $58.0\pm 0.8$ & $61.1\pm 4.1$ & $67.6$ \\ 
        & Walker & $81.3\pm0.3$ & $77.6\pm0.9$ & $79.2\pm0.6$ & $82.2\pm0.0$ & $75.4\pm1.0$ & $74.0$ \\ 
        \cmidrule{2-8}
        & Average & $60.2$ & $57.6$ & $59.1$ & $58.0$ & $58.4$ & $61.4$ \\ 
    
    \midrule
    \multirow{3}{*}{\shortstack{Medium\\Expert }} 
        & HalfCheetah & $47.4\pm0.5$ & $40.1\pm1.0$ & $39.3\pm0.1$ & $35.4\pm1.0$ & $36.4\pm0.2$ & $82.2$ \\ 
        & Hopper & $49.4\pm3.1$ & $41.0\pm1.2$ & $71.5\pm3.3$ & $47.7\pm1.8$ & $91.5\pm1.7$ & $109.1$ \\ 
        & Walker & $101.9\pm0$ & $94.9\pm2.2$ & $98.6\pm 5.5$& $83.6\pm2.2$ & $106.9\pm1.0$ & $107.9$ \\ 
        \cmidrule{2-8}
        & Average & $66.2$ & $58.6$ & $69.8$ & $55.6$ & $78.2$ & $99.7$ \\     

    
    
    \bottomrule
  \end{tabular}
\end{table*}

\subsection{Merging with language pre-training}
\label{sec:lm}


We may obtain better results if we merge two (or more) models which share a common initialization. Additionally, if models are encouraged to stay close to this original initialization, then we may also see better merging results. 



We train using Equation \ref{eq:chibi}, reporting several variants that all decay $\lambda_1$ to $0.0$ after 5000 following \cite{wiki}, this includes using language co-training (ChibiT$_\mathcal{CO}$: $\lambda_2 = 1, \lambda_3=0$), using $l_2$ regularization to the initialization (ChibiT$_\mathcal{R}$: $\lambda_2=0, \lambda_3=1$), or no co-training or regularization (ChibiT: $\lambda_2 = 0, \lambda_3=0$).

We train models for HalfCheetah, Hopper, and Walker medium and medium-expert tasks and evaluate \textit{Merge-Freeze-Finetune} (MFF), merging across all three models, instead of between pairs. We also merge over attention and MLP layers. We display results in Table \ref{table:lm-merge}.  In addition to language-initialized models, we report several baselines. This includes repeating the MFF process but with randomly initialized models as in previous sections (Random). We also train a multi-task baseline (Multi), which simultaneously trains a decision transformer on all three tasks. We also report the results of training individual DTs for each task (Single). First, we notice a performance gap between the multi-task and single-task models, as multi-task models are the same size as single-task models, but must learn all tasks. Our goal is to obtain similar performance to the multi-task model but without simultaneous, multi-task training. We find that this is successful on MuJoCo medium, with ChibiT and ChibiT$_\mathcal{R}$ exceeding the performance of the multi-task model, and coming close to single-task performance. Because merged models share attention and MLP parameters, we obtain a model with just $100.52\%$ the transformer parameters of a single task model, for all three tasks, as each transformer block for each task only has unique layer-normalization layers, but shares all other parameters. On medium-expert, we see a larger gap between both our merged, and multi-task models to the single-task models. We continue by discussing two complementary methods for closing this gap on the medium-expert benchmark.

\begin{table*}[h]
  \centering
  \small
    \caption{\textbf{Merging with increased model size and Fisher information} with \textit{Merge-Freeze-Finetune} (MFF) over all three models at once on D4RL medium-expert, extending Table \ref{table:lm-merge}. We list the score of the model after performing MFF, followed by the initial score after training for a specific task, before merging (MFF $/$ Initial). We evaluate merging with both {ChibiT}$_\mathcal{R}$ and GPT2$_\mathcal{R}$ initializations. }
  \label{table:lm-merge-improved}
  \begin{tabular}{cllllll}
    \toprule
    
    \multirow{2}{*}{\textbf{Dataset}}     &  \multirow{2}{*}{\textbf{Environment}} & \multicolumn{2}{c}{\textbf{ChibiT}$_\mathcal{R}$} & \multicolumn{2}{c}{\textbf{GPT2$_\mathcal{R}$}} & \multirow{2}{*}{\textbf{Single}} \\
    \cmidrule(lr){3-4}  \cmidrule(lr){5-6}
    & & \multicolumn{1}{c}{Direct} & \multicolumn{1}{c}{Fisher} & \multicolumn{1}{c}{Direct} & \multicolumn{1}{c}{Fisher} & \\
    \midrule
    \multirow{3}{*}{\shortstack{Medium\\Expert}} 
        & HalfCheetah & $39.3 ~/~ 89.5$ & $42.6 ~/~ 89.5$ & $84 ~/~ 89.9$ & $87.2 ~/~ 89.9$ & $82.2$ \\ 
        & Hopper & $71.5 ~/~ 106.9$ & $77.2 ~/~ 106.9$ & $84.1 ~/~ 106.6$ & $88.3 ~/~ 106.6$ & $109.1$ \\ 
        & Walker & $98.6 ~/~ 109.3$ & $103.3 ~/~ 109.3$ & $109 ~/~ 107.8$ & $109.4 ~/~ 107.8$ & $107.9$  \\ 
        \cmidrule{2-7}
        & Average & $69.8 ~/~ 101.9$ & $74.4 ~/~101.9$ & $92.4 ~/~ 101.4$ & $95.0 ~/~ 101.4$ & $99.7$  \\ 
    
    \bottomrule
  \end{tabular}
\end{table*}

\subsection{Fisher merging and larger pre-trained initializations}
\label{sec:ext}
In this section, we jointly consider two sources of improvement. First, we consider repeating our procedure of {ChibiT}$_\mathcal{R}$ in Section \ref{sec:lm} but with \textbf{larger models}, specifically GPT2 small, with 84M transformer parameters. Additionally, we consider \textbf{Fisher merging}, to replace directly averaging. We evaluate these two improvements on medium-expert, as shown in Table \ref{table:lm-merge-improved}, reporting average results over two instances of MFF. We can see that combining both larger model initializations and Fisher merging results in a multi-task model that is achieved without any centralized training, sharing transformer attention and MLP parameters over all tasks, and attaining an average D4RL   normalized score of $94.97$, close to the score of training separate individual Decision Transformers ($99.73$). Additionally, while training with larger models results in similar initial performance, we see significantly smaller reduction in performance after merging. 

\label{sec:multi-finetune}
\begin{table}[h]
      \centering
      \caption{\textbf{Merging with finetuned multi-task models}. We display performance of initial multi-task model, and then after finetuning on expert data for each task separately. Next, we show merging the resulting models. We repeat initial training twice, and merging and evaluating performance twice, reporting the mean and standard error over four evaluations.  }
  \label{table:multi-merge}

    \begin{tabular}{llll}
        \toprule    
        Model(s) & \shortstack{Hopper} & \shortstack{Cheetah} & Walker  \\ 
        \midrule
        Multi-Task Medium & $72 \pm 1.9$ & $39.9\pm0.1$ & $74.9\pm0.9$ \\ 
        Finetuned Experts & $108.3\pm 1.4$ & $70.5\pm 0.3$ & $108.8\pm0.2$ \\
        \midrule
        Direct Merge & $60.2\pm3.2$ & $43.7\pm0.9$ & $70.4\pm7.8$ \\ 
        Fisher Merge & $\mathbf{84.3\pm2.8}$ & $\mathbf{45.5\pm0.6}$ & $\mathbf{99\pm1.2}$ \\ 

        \bottomrule
    \end{tabular}
\end{table}

\subsection{Merging finetuned multi-task models}

We also examine if we can have settings where we can improve models, through merging, but without additional finetuning after merging. We consider multi-task models, and as report a gap between small multi-task and single-task models in Section \ref{sec:lm}, we also use a larger model with embedding dimension of 512, and 4 heads, and 3 layers for this experiment. We begin by training a multi-task model on the medium datasets. After training this model, we independently finetune the model on expert data for each task with initialization regularization as in Section \ref{sec:lm}, resulting in three separate finetuned models. We then merge all transformer parameters of the resulting finetuned models, to obtain one multi-task model. We evaluate both regular merging, and Fisher merging in Table \ref{table:multi-merge}. When directly averaging the weights of finetuned expert models, we see performance decrease on Walker and Hopper, and slightly improve from the medium model with HalfCheetah. However, by performing Fisher merging, we improve considerably on Hopper and Walker, with still a slight improvement on HalfCheetah. While a gap still exists between our merged multi-task model, and the maximum performance achieved by each separate finetuned model, it is promising to find that we can have settings where merging improves performance on all tasks, even without MFF, as used in previous sections. 





\section{Discussion \& Conclusion}

In this work, we studied merging Decision Transformers trained on MuJoCo locomotion problems. We found that it is possible to merge randomly initialized models, leading us to analyze the role of attention in DTs, use merging to obtain or improve multi-task models, and investigate using additional finetuning, common initializations, and utilizing Fisher information. 

While we can come close to performance of individually trained policies, further work could explore merging with reduced penalty. Additionally, we believe follow-ups of our work could see if our findings more generally hold beyond merging policies locomotion tasks. This includes other multi-task settings \cite{multi-game-dt} with DTs, testing other kinds of pre-training \cite{Pre-trainingForRobotsNewtasks, perception-action-PACT} besides language, or merging other kinds of models in the RL settings which incorporate pre-training and finetuning \cite{multi-q, atari-pretraining-finetune, cross-task-transfer-modelRL}. In general, we view a broader impact of merging as providing a possible direction for more accessibly creating multi-task DTs or generalist models. 






{
\bibliographystyle{plain}
\bibliography{refs} 
}

\end{document}